\documentclass[11pt,a4paper]{article}
\usepackage{natbib}
\usepackage{times}
\usepackage{latexsym}
\usepackage{amsmath}
\usepackage{multirow}
\usepackage{url}

\usepackage{amsfonts}
\usepackage{paralist}
\usepackage{graphicx}
\usepackage{tikz}
\usepackage{linguex}
\usepackage{subcaption}
\usepackage{booktabs}
\usepackage[utf8]{inputenc}
\usepackage[affil-it]{authblk}

\usetikzlibrary{arrows,automata,shapes}
\title{Fast and unsupervised methods for multilingual cognate clustering}

\author{Taraka Rama, Johannes Wahle, Pavel Sofroniev, and Gerhard Jäger}
\affil{University of Tübingen}

\date{}

\begin{document}
\maketitle
\begin{abstract}
In this paper we explore the use of unsupervised methods for detecting cognates 
in multilingual word lists. We use online EM to train sound segment similarity weights for 
computing similarity between two words. We tested our 
online systems on geographically spread sixteen different language groups of 
the world and show that the Online PMI system (Pointwise Mutual Information) 
outperforms a HMM based system and two linguistically motivated systems: LexStat 
and ALINE. Our results suggest that a PMI system trained in an online fashion can 
be used by historical linguists for fast and accurate identification of cognates 
in not so well-studied language families.
\end{abstract}

\section{Introduction}
Cognates are genetically related words that can be traced to a common word in a language that 
is no longer spoken. For example, English 
\emph{nail} and German \emph{nagel} are cognates with each other which can be traced back to the 
stage of Proto-Indo-European *$h_3enog^h-$. Accurate identification of cognates is important for 
inferring the internal structure of a language family.

Recent years has seen an surge in the number of publications in the field of computational 
historical linguistics due to the availability of word 
lists for large number of languages of the world 
\citep{brown2013sound}\footnote{Known as 
Automated Similarity Judgment Program (ASJP). \url{http://asjp.clld.org/}} and cognate databases 
for Austronesian \citep{greenhill2009austronesian} and Indo-European 
 \citep{bouckaert2012mapping}. 

The availability of word lists (without cognate judgments) has allowed scholars like 
\citet{rama2013bchap} and \citet{Jaeger2015} to experiment with different weighted string 
similarity measures 
for the purpose of inferring the family trees of world's languages, without explicit cognate 
identification. On the other hand, \citet{list2012lexstat} proposed a cognate clustering system 
that 
combines handcrafted weighted string similarity measures and permutation tests for the purpose of 
automated cognate identification. In a different approach, \citet{Hauer2011} 
experimented with linear classifiers like SVMs for the purpose of identifying cognate clusters. 
Finally, \citet{rama2015automatic} use string kernel inspired features for training a SVM linear 
classifier for pair-wise cognate identification. As noted by \citet{Hauer2011}, availability of a 
reliable multilingual cognate identification system can be used to supply the cognate judgments as 
an input to the phylogenetic inference algorithms introduced by \citet{gray2003language} and 
reconstruction methods of \citet{bouchardcote2013}.\footnote{The cognate clustering system in 
\citet{bouchardcote2013} requires the tree structure of the language family to be know beforehand. 
This is not a practical assumption since the tree structure of many language families of the world 
is not known beforehand.}


The phylogenetic inference methods require cognate judgments which are only available for a small 
number of well-studied language families such as Indo-European and Austronesian. For instance, the 
ASJP database 
provides Swadesh word lists (of length $40$ which are resistant to lexical replacement and 
borrowing) 
transcribed in a uniform format for more than $60\%$ of the world's languages. However, the cognacy 
judgments are only available for a subset of language families. An example of such a word list is 
given in table \ref{tab:Datasample}.
\begin{table}[!ht]
  \centering
    \begin{tabular}{|l|cccc|}
   \hline
       & ALL & AND & ANIMAL & $\ldots$\\
          \hline
    English & ol & End & Enim3l & $\ldots$\\
    German & al3 & unt & tia & $\ldots$\\
    French & tu & e  & animal & $\ldots$\\
    Spanish & to8o & i  & animal & $\ldots$\\
    Swedish & ala & ok & y3r & $\ldots$\\
       \hline
    \end{tabular}%
    \caption{Example of a word list for five languages belonging to Germanic (English, German, and 
Swedish) and Romance (Spanish and French) subfamilies transcribed in ASJP alphabet.}
\label{tab:Datasample}
\end{table}

The task at hand is to automatically cluster words that show genealogical relationship. This is 
achieved by computing similarities between all the word pairs belonging to a meaning and then 
supplying the resulting distance matrix as an input to a clustering algorithm. The clustering 
algorithm groups the words into clusters by optimizing a similarity criterion. The similarity 
between a word pair can be computed using supervised approaches \citep{Hauer2011} or by using 
sequence alignment algorithms such as Needleman-Wunsch \citep{NW} or Levenshtein distance 
\citep{Levenshtein1966}.

  %

In dialectometry, \citet{WielingNerbonne2007} compared Pair Hidden 
Markov Model (PHMM) \citep{MackayKondrak2005} and pointwise mutual information (PMI)
\citep{Church:90} weighted Levenshtein distance for Dutch dialect comparison. In 
historical linguistics, \citet{jager2013phylogenetic} developed a PMI 
based method for computing the string similarity using the ASJP database. In this paper, we apply 
\emph{online} algorithms to train our PMI and PHMM systems for the purpose of computing word 
similarity.

We train our PHMM and PMI systems in different settings and test it on sixteen different families 
of the world. Our results show that online training can perform better 
than a linguistically well-informed system known as LexStat \citep{list2012lexstat}. Also, the 
online algorithms allow our systems to be trained in few minutes and give similar accuracies as 
the batch trained systems of \citet{jager2013phylogenetic}.

The paper is organized as follows. We discuss the relevant work in section 
\ref{sec:relwork}. We describe the PMI and PHMM models in section 
\ref{sec:models}. 
The Online EM procedure is described in section \ref{sec:oneem}. We describe the 
clustering algorithm in section \ref{sec:cluster}. We discuss the experimental settings and 
motivation behind our choices in section \ref{sec:exps}. We present and discuss the results of our 
experiments in section \ref{sec:results}. We discuss the effect of different model parameters in 
section \ref{sec:erranal}. Finally, we conclude the paper in section \ref{sec:concl}.

\section{Related work}\label{sec:relwork}

\citet{kondrak2000new} introduced a dynamic programming algorithm for 
computing the similarity between two sequences based on articulatory phonetic 
features determined by \citet{Ladefoged1975}. The author evaluated his 
algorithm on a list of English-Latin cognates. In this paper, we evaluate on 
the Indo-European dataset consisting of English and Latin.

\citet{Hauer2011} trained a linear SVM on word similarity features and use the 
SVM model to assign 
a similarity score to the word pair. For each meaning, a word pair distance 
matrix is computed and supplied to the average linkage clustering algorithm for 
inferring cognate clusters. The authors observe that the SVM trained system 
performs better than a baseline that judges the similarity of two words based on the identity of 
the 
first two consonants.

\citet{list2012lexstat} introduced a system known as LexStat (described in section 
\ref{sec:exps}) that is sensitive to segment similarities and chance similarities due to borrowing 
or semantic shift.
 The author tests this system on a number of small-sized 
(consisting of less than $20$ languages) datasets for the purpose of cognate identification and 
reports that the system performs better than Levenshtein distance.

In a recent paper, \citet{list-lopez-bapteste:2016:P16-2} explore the use of InfoMap 
\citep{Rosvalletal2008} for the detection of partial cognates in subgroups of 
Sino-Tibetan language 
family. The authors compare the performance of average linkage 
clustering against InfoMap and find that InfoMap performs better than average 
linkage 
clustering. 

The above listed works test similar datasets using different experimental settings. For 
instance, \citet{Hauer2011} trained and tested on a subset of language families that were 
provided by \citet{wichmann2013languages}. At the same time, to the best of our knowledge, the 
LexStat system has not been evaluated on all the available language families. 
Moreover, the PMI-LANG 
\citep{jager2013phylogenetic}, has not been evaluated at the task of 
unsupervised cognate clustering.

\section{Models}\label{sec:models}
In this section, we briefly describe the PMI weighted Needleman-Wunsch algorithm and Pair Hidden 
Markov Model (PHMM). 

\subsection{PMI-weighted alignment}
The vanilla Needleman-Wunsch (VNW) algorithm is the similarity counterpart of 
the
Levenshtein distance. It maximizes similarity whereas Levenshtein distance 
minimizes the
distance.  In VNW, a character or sound segment match increases the similarity 
by $1$ and
a character mismatch has a weight of $-1$. In contrast to Levenshtein distance 
which
treats insertion, deletion, and substitution equally, VNW introduces a gap 
opening
(deletion operation) penalty parameter that has to be set separately. A second 
parameter
known as gap extension penalty has lesser or equal penalty than the gap opening 
parameter
and models the fact that deletions occur in chunks 
\citet{jager2013phylogenetic}.

VNW is not sensitive to segment pairs, but a realistic algorithm should assign 
higher
similarity score to sound correspondences such as /l/ $\sim$ /r/ than the sound
correspondences /p/ $\sim$ /r/. The \emph{weighted} Needleman-Wunsch algorithm 
requires a
\emph{similarity score} for each pair of segments, and it finds the alignment(s) 
betwen
two input strings maximizing the sum of the pairwise similiarities of matched 
segment
pairs.

In computational historical linguistics, similarity between two segments is 
estimated
using PMI. The PMI score for two sounds $i$ and $j$ is defined as followed:
\begin{equation}\label{eq:PMI} 
\textnormal{PMI}(i,j) = \log \frac{p(i,j)}{q(i)\cdot q(j)}
\end{equation}
where, $p(i,j)$ is the probability of $i, j$ being matched in a pair of cognate 
words,
whereas, $q(i)$ is the probability that an arbitrarily chosen segment in an 
arbitrarily
chosen word equals $i$. A positive PMI value between $i$ and $j$ indicates that 
the
probability of $i$ being aligned with $j$ in a pair of cognates is higher than 
what would
be expected by chance. Conversely, a negative PMI value indicates that an 
alignment of $i$
with $j$ is more likely the result of chance than of shared inheritance.

We estimated PMI scores from raw data, largely following the method described in
\citet{jager2013phylogenetic}.

The whole training procedure can be described as follows:
\begin{enumerate}
\item Extract a set of word pairs that are probably cognate using a 
suitable heuristics. In this paper, we treat all word pairs belonging to the 
same meaning with a length normalized Levenshtein distance (LDN) below $0.5$ as 
\emph{probable cognates}.\footnote{We experimented with LDN cutoffs of $0.25$ and $0.75$ and found 
that the results are best for a cutoff of $0.5$}
\item Align the list of probable cognates using the vanilla Needleman-Wunsch 
algorithm.
\item Extract aligned segment pairs and compute the PMI value for a segment pair 
using equation \ref{eq:PMI} and estimating probabilities as relative 
frequencies.
\item Generate a new set of aligments using Needleman-Wunsch algorithm and the 
segment weights learned from step 2. For the gap penalties we used the values 
proposed in \citet{jager2013phylogenetic}.
 \item We iterate between step 2 and 3 until the average similarity between the 
two iterations does not change.
\end{enumerate}

This procedure yields a PMI based similarity score for each word pair. We 
convert the similarity 
score $x$ into a distance score using the sigmoid transformation: $1.0 - 
(1+exp(-x))^{-1}$ that 
converts the PMI similarity score into the range of $[0,1]$. 

\subsection{Pair Hidden Markov Model}
Pair Hidden Markov Model was first proposed in the context of computational biology as a tool for
the comparison of DNA or protein sequences \citep{Durbinetal}.

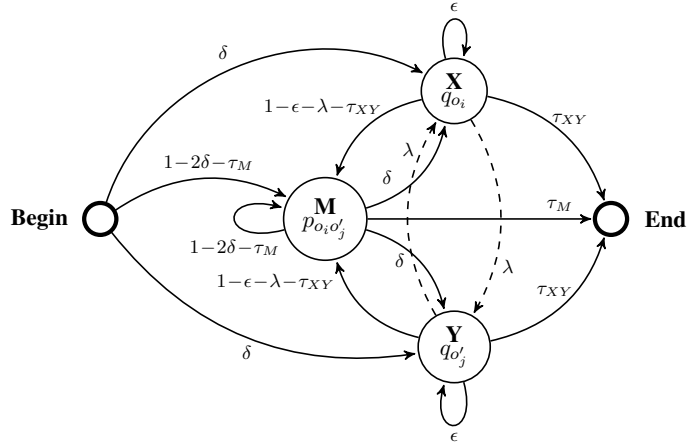
\begin{figure}[t]
\centering
\begin{tikzpicture}[->,>=stealth',shorten >=1pt,auto,node distance=3cm,
                    semithick, scale=0.8, every node/.style={transform shape}]
  \tikzstyle{every state}=[fill=none,draw=black,text=black, align=center]

  \node[state]         (A)                                {\textbf{M}\\[-0.7em]$p_{o_{i}o'_{j}}$};
  \node[state]         (B) [above right of=A] {\textbf{X}\\[-0.7em]$q_{o_i}$};
  \node[state]         (C) [below right of=A] {\textbf{Y}\\[-0.7em]$q_{o'_j}$};
  \node[state]         (D) [left of=A, node distance = 3.7cm, label={[xshift = -10mm, 
yshift=-6mm]\textbf{Begin}}, scale=0.55, ultra thick]              { };
  \node[state]         (E) [right of=A, node distance = 4.7	cm, label={[xshift = 9mm, 
yshift=-5.5mm]\textbf{End}}, scale=0.55, ultra thick]              { };
 \node (dummy2) [below left of = C, node distance=2.7cm, xshift=10mm] {};

  \path (A) edge  [bend right]            node[pos=0.3] {\footnotesize{$\delta$}} (B)
            edge    [bend left	]          node[below, pos=0.3] {\footnotesize{$\delta$}} (C)
            edge node[pos=0.85]{\footnotesize{$\tau_M$}} (E)
            edge [loop left] node[below, yshift=-2mm]{\footnotesize{$1\!-\!2\delta\!-\!\tau_M$}} (A)
        (B) edge [loop above] node {\footnotesize{$\epsilon$}} (B)
            edge    [bend left]          node[xshift=-3mm] {\footnotesize{$\tau_{XY}$}} (E)
            edge [bend right] node[above, pos=0.45, 
xshift=-8mm]{\footnotesize{$1\!-\!\epsilon\!-\!\lambda\!-\!\tau_{XY}$}} (A)
            edge [bend left, dashed] node[pos=0.75, right]{\footnotesize{$\lambda$}} (C)
        (C) edge   [bend right]           node[xshift=3mm] {\footnotesize{$\tau_{XY}$}} (E)
            edge [loop below]  node {\footnotesize{$\epsilon$}} (C)
            edge [bend left, dashed] node[pos=0.85, left]{\footnotesize{$\lambda$}} (B)
            edge [bend left] node[above, pos=0.65, 
xshift=-13mm]{\footnotesize{$1\!-\!\epsilon\!-\!\lambda\!-\!\tau_{XY}$}} (A)
        (D) edge [bend left = 50] node {\footnotesize{$\delta$}} (B)
            edge    [bend left]          node[pos=0.55] {\footnotesize{$1\!-\!2\delta\!-\!\tau_M$}} 
(A)
            edge [bend right] node[below]{\footnotesize{$\delta$}} (C);

\end{tikzpicture}
\caption{Pair Hidden Markov model as proposed by Mackay \& Kondrak 
{\protect\citep{MackayKondrak2005}}. The states of the model are depicted as the circles. The 
arrows 
show the possible transitions between the states. $<\delta, \epsilon, \lambda, \tau_M, \tau_{XY}>$ 
represent the transition probabilities.}
\label{fig:PairHMMkondrak}
\end{figure}

A Pair Hidden Markov Model (PHMM) uses two output streams, instead of a single output stream;
one for each of the two sequences being aligned. In its simplest version, a PHMM consists of 
five states. A \emph{begin} state, an \emph{end} state, a match state (\textbf{M}) that emits pairs 
of symbols, a 
deletion state (\textbf{X}) that emits a symbol in the first string and a gap in the second string; 
and an insertion state (\textbf{Y}) that emits a gap in the first and a symbol in the second string 
(cf. figure \ref{fig:PairHMMkondrak}). 

The PHMMs, as used in historical linguistics, differ from its biological counterpart in 
the following regard:
\begin{itemize}
 \item The historical linguistic PHMM allows a transition between the states \textbf{X} and 
\textbf{Y} (dashed line, figure \ref{fig:PairHMMkondrak}). An alignment of Italian \emph{due} and 
Spanish \emph{dos} `two' cannot be generated by a PHMM without the transition between \textbf{X} 
and \textbf{Y} \citep{MackayKondrak2005}.
\begin{center}
d u e -\\
d o - s\\
\end{center}
\item Another difference between the biological and the linguistic PHMM is the split of 
the parameter for the transition into the end state. Whilst the original version only has one 
parameter for this purpose, the linguistic PHMM makes use of two different probabilities 
$\tau_M$ and $\tau_{XY}$. This split of parameters enables the model to distinguish between the 
match state (\textbf{M}) being the final emitting state or any of the gap states 
(\textbf{X},\textbf{Y}) (see figure \ref{fig:PairHMMkondrak}). This modification preserves the 
symmetry of the model, while allowing a little bit more freedom.
\end{itemize}

The PHMMs are trained using Baum-Welch expectation maximization algorithm \citep{Durbinetal}. 
The best alignment between two sequences $x$ and $y$ is determined by using the Viterbi algorithm.

The probability of two sequences $x$ and $y$ of lengths $m$ and $n$ respectively evolving 
independently under a null model $R$ is given by the following equation \ref{eq:randommodel}.
\begin{equation}
\label{eq:randommodel}
P(x,y|R) = \iota^2(1-\iota)^{n+m} \prod \limits_{i=1}^{n}f_{x_i} \prod \limits_{j=1}^{m}f_{y_j},
\end{equation}
with $f_{x_i}$ is the equilibrium frequency of the sound at position $i$ in sequence $x$ where, 
$\iota = \frac{1}{\frac{m+n}{2} + 1}.$

The probability of relatedness between $x$ and $y$ is computed as the logarithmic ratio of the 
probability scores $P(x,y|\mu)$ and $P(x,y|R)$, where $\mu$ is the trained model and $R$ is the 
null model.


We employ the same sigmoid transformation, as in PMI, to convert the similarity score 
(computed under a PHMM) to a distance score.

\section{Online EM}\label{sec:oneem}
The Expectation Maximization algorithm (EM) is widely used in computational linguistics for the 
purpose of word alignment, document classification, and word segmentation. The EM algorithm starts 
with an initial setting of model parameters and uses that model parameters to realign words in a 
sentence pair. The model parameters are reestimated using the word alignments obtained from the 
previous iteration. The EM algorithm reestimates the model parameters after each full scan of the 
training data.

\citet{LiangKlein2009} observe that batch training procedure 
can lead to slow convergence. As a matter of 
fact, \citet{jager2013phylogenetic} trains his PMI system using 
the standard EM (also known as batch EM) which updates the parameters in a PMI scoring matrix only 
after aligning all the 
word pairs. In contrast, Online EM \citep{LiangKlein2009}, updates the model parameters after 
aligning a subset of word pairs (also known as minibatch in online learning literature).

The Online EM algorithm combines the parameters estimated ($s$) from the current update step $k$ 
with the previous parameters $\theta_{k-1}$ using the following equation:

\begin{equation}
 \theta_k = (1-\eta_k) \theta_{k-1} + \eta_k s
\end{equation}
where $\eta_k$ is defined as: $\eta_k = (k+2)^{-\alpha}$.

In the case of PMI, $\theta$ constitutes the PMI scores for all segment pairs. The parameter 
$\eta_k$ determines how fast to forget or remember the updates from the previous steps. The 
$\alpha$ 
parameter is in the range of $0.5 \le \alpha \le 1$. A smaller $\alpha$ implies a large update to 
the model parameters. The parameter $k$ is related to minibatch parameter ($m$; $m = \lceil D/k 
\rceil$; where, $D$ is the size of training data) and determines the number of updates to be 
performed. The setting $k=1$ recovers the batch EM whereas, when $k=D$, implies 
an update for each sample in the training data.

\section{Clustering algorithm}\label{sec:cluster}


The InfoMap clustering method is an information theoretic approach to detect 
community structure within a connected network. The method uses random walks on a network as a 
proxy for information flow to detect communities, i.e., clusters, without the need for a threshold. 
A community is a group of nodes with more edges 
connecting the nodes within the community than connecting them with nodes outside the 
community \citep{NewmanGirvan2004}.

In our case, a community refers to the words which are cognate 
and have higher edge weights between them. The idea behind the algorithm is that the random walk is 
statistically more likely to spend a long period of time within a community than switching 
communities due to the nature of the network.

A pair-wise distance matrix is a complete weighted graph and any edge that has a weight $< 0.5$ and 
a PMI score $< 0$ (due to the sigmoid-based distance transformation). Due to the PMI score's 
definition, a PMI score $< 0$ implies that the words might not be cognate. We use this property to 
construct a non-complete graph and supply the resulting network as an input to the InfoMap 
algorithm. 

\section{Experiments}\label{sec:exps}
In this section, we describe the experimental settings, datasets, evaluation measures, and the 
comparing systems: Baseline, ALINE, PMI-LANG, and LexStat.
 
\subsection{Hyperparameters of Online EM}
We determine the best setting of $m$ and $\alpha$ parameter by searching for $m$ in the range of $m 
= 2^s$ where $s \in [5, 15]$; and, $\alpha \in [0.5, 1.0]$ with a step size of $0.05$. We fix the 
gap opening and gap extension penalties to $-2.5$ and $-1.75$.

\subsection{Datasets}
\subsubsection{Indo-European database}
The Indo-European Lexical database (IELex) was 
created by \citet{dyen1992indoeuropean} and curated by Michael 
Dunn.\footnote{\url{http://ielex.mpi.nl/}} The IELex database is not 
transcribed in uniform IPA and retains many forms transcribed in 
the Romanized IPA format of \citet{dyen1992indoeuropean}. We cleaned the IELex database of any 
non-IPA-like transcriptions and converted the cleaned subset of the database 
into ASJP format.

\subsubsection{Austronesian vocabulary database} The Austronesian Vocabulary 
Database (ABVD) \citep{greenhill2009austronesian} has word lists for $210$ Swadesh concepts and 
$378$ 
languages.\footnote{\url{http://language.psy.auckland.ac.nz/austronesian/}} The database 
does not have transcriptions in a uniform IPA format. We 
removed all symbols that do not appear in the standard IPA and converted the lexical items to ASJP 
format. For comparison purpose, we use randomly selected $100$ languages' dataset in this 
paper.\footnote{LexStat takes many hours to run on a dataset of $100$ languages.}

\subsubsection{Short word lists with cognacy judgments:} \citet{wichmann2013languages} and 
\citet{List2014d} compiled cognacy wordlists for subsets of families from 
various scholarly sources such as comparative handbooks and historical linguistics' articles. 
The details of different databases is given in table \ref{tab:chardataset}.


\begin{table}[!ht]
\small
\centering
    \begin{tabular}{lcccc}\hline
    Family   & NOM & NOL &  AveCC & AveWC \\\hline
    Austronesian & 210 & 100 & 20.2142 & 4.1143 \\
    Afrasian & 40 & 21 & 9.5 & 2.6868 \\
    Bai dialects & 110 & 9  & 2.5909 & 6.0166 \\
    Chinese dialects & 179 & 18 & 6.8771 & 5.2635 \\
    Huon & 84 & 14 & 6.3929 & 2.7672 \\
    Indo-European & 207 & 52 & 12.2126 & 7.3461 \\
    Japanese dialects & 200 & 10 & 2.3 & 6.1373 \\
    Kadai & 40 & 12 & 3.225 & 5.0027 \\
    Kamasau & 36 & 8  & 1.6667 & 5.3981 \\
    Lolo-Burmese & 40 & 15 & 2.625 & 7.3121 \\
    Mayan & 100 & 30 & 8.58 & 6.1521 \\
    Miao-Yao & 39 & 6  & 1.8974 & 3.9667 \\
    Mixe-Zoque & 100 & 10 & 3  & 4.6535 \\
    Mon-Khmer & 100 & 16 & 7.75 & 2.7956 \\
    ObUgrian & 110 & 21 & 2.2 & 11.8162 \\
    Tujia & 109 & 5  & 1.6422 & 3.3792 \\\hline
    \end{tabular}%
    \caption{Number of languages (NOL), Number of meanings (NOM), Average number of cognate classes 
per meaning (AveCC), and Average number of words per cognate class (AveWC).}
  \label{tab:chardataset}%
\end{table}%


\subsection{Evaluation Measures}
We evaluate the results of clustering analysis using B-cubed F-score \citep{amigo2009comparison}. 
The B-cubed scores are defined for each word belonging to a meaning as followed. The precision 
for a word is defined as the ratio between the number of cognates in its cluster to the total 
number of words in its cluster. The recall for a word is defined as the ratio between the number 
of cognates in its cluster to the total number of expert labeled cognates. The B-cubed precision 
and recall are defined as the average of the words' precision and recall across all the clusters. 
Finally, the B-cubed F-score for a meaning, is computed as the harmonic mean of the average items' 
precision and recall. The Averaged B-cubed F-score for the whole dataset is computed as the average 
of the B-cubed F-scores across all the meanings.

\citet{amigo2009comparison} show that the B-cubed F-score satisfies four formal constraints known 
as cluster homogeneity, cluster completeness, rag bag (robustness to misplacement of a true 
singleton item), and robustness to variation in cluster size. The authors show that cluster 
evaluation measures based on entropy such as Mutual Information and V-measure 
\citep{rosenberg2007v} and Rand index do not satisfy the four constraints. Both 
\citet{Hauer2011} and \citet{list-lopez-bapteste:2016:P16-2} use B-cubed 
F-scores to evaluate their cognate clustering systems.

\subsection{Comparing systems}
\textbf{Baseline} We adopt length normalized Levenshtein distance as the baseline in our 
experiments.

\subsubsection{ALINE}
ALINE is a sequence alignment system designed 
by \citet{kondrak2000new} for computing similarity between two words by 
decomposing phonemes into multivalued and binary phonetic features. Each phoneme is decomposed into 
multivalued features such as place and manner for consonants; height 
and backness for vowels. Multivalued features take values on a continous scale 
ranging from $[0, 1]$ and the values represent the distance between the sources 
of articulation. Binary valued features consist of nasal, voicing, aspirated, and retroflex. 

Each feature is weighed by a \emph{salience} value that is determined manually. The similarity 
score between two sequences is computed as the sum of the aligned sound 
segments. Following \citet{downey2008computational}, we convert ALINE's 
similarity score $s_{ab}$ between two words $a,b$  is converted to a distance score 
based on the following formula: 
$1.0-\frac{2.0*s_{ab}}{s_{aa}+s_{bb}}$.\footnote{We use the Python 
implementation provided by \citet{huff2011positing} which is available at 
\url{https://sourceforge.net/projects/pyaline/}.}

\subsubsection{PMI-LANG}
\citet{jager2013phylogenetic} developed a system that 
learns PMI sound matrices to optimize a criterion designed to optimize language 
relatedness. The core idea is to tie up word similarity to language similarity 
such that close languages such as English/German tend to have more 
similarity than English/Hindi. The language similarity function amounts 
to maximizing similarity between probable cognates to learn a PMI score matrix. 
\citet{jager2013phylogenetic} applies the learned PMI score matrix to infer 
phylogenetic trees of language families. However, the learned PMI score matrix 
has not been applied for cognate clustering.

\subsubsection{LexStat}
LexStat \citep{list2012lexstat} is part of LingPy \citep{List2016e} 
library offering state-of-the-art alignment algorithms for aligning word pairs 
and clustering them into cognate sets. We describe the workflow of LexStat 
system below:
\begin{enumerate}
 \item LexStat uses a hand-crafted sound segment matrix, $h$, to align and score the word pairs for 
each 
meaning. Let a segment pair $i,j$'s similarity be given as $h_{ij}$.
\item For each language pair, $l_1, l_2$ the word pairs belonging to the same meaning are aligned. 
The frequency of a segment pair $i,j$ belonging to the same meaning is given as $a_{ij}$.
\item For $l_1, l_2$, the words belonging to one of the language is shuffled and realigned using 
Needleman-Wunsch algorithm. This procedure is repeated for all language pairs for $100$ times. The 
average frequency of a segment pair $i,j$ from the reshuffling step is given as $e_{ij}$.
\item All the parameters $h, a, e$ are combined according to the following formula to give a new 
segment similarity score $s_{ij}$ where, $w_1+w_2=1$.
\begin{equation}
 s_{ij} = 2*w_1 log \frac{a_{ij}}{e_{ij}} + w_2h_{ij}
\end{equation}
\item The weights $s_{ij}$ are then used to score word pairs and cluster words in a meaning.
\end{enumerate}

The intuition behind step 3 is to reduce the effect of chance 
similarities between the sound segments that can obscure real genetic sound 
correspondences.\footnote{We obtained the code from \url{https://github.com/lingpy}. We convert the 
LexStat similarity scores into distance scores using the same formula as ALINE.} We supply the word 
distances from all the above systems as input to InfoMap to infer cognate 
clusters.


\section{Results}\label{sec:results}
In this section, we present the results of our experiments. We perform two sets of experiments by 
training with different datasets which are described below.
\begin{table*}[!ht]
  \centering
\scalebox{0.6}{
    \begin{tabular}{l|c||c|c|c|c|c|c|c}
    \hline
     Family  &  LDN  &  PMI-LANG  &   Batch PMI &  Online PMI  &  Batch PHMM 
 
&  Online PHMM  &  LexStat  &  ALINE \\\hline
Austronesian  & 0.7175 & 0.7355 & 0.6539 &  \textbf{0.7364}  & 0.6224 & 0.6709 & 
0.7173 & 0.5321\\
Afrasian  & 0.7993 & 0.8133 & 0.7496 &  \textbf{0.8392}  & 0.7213 & 0.7044 &  --  & 0.6442\\
Bai dialects  & 0.8348 & 0.8766 & 0.8716 &  \textbf{0.8774}  & 0.8741 & 0.8639 & 0.8417 & 0.8462\\
Chinese dialects  & 0.7687 & 0.7521 & 0.7217 & 0.7803 & 0.7455 & 0.7396 &  \textbf{0.7815}  & 
0.6651\\
Huon  & 0.8536 & 0.8556 & 0.7518 & 0.8775 & 0.7612 & 0.7437 &  --  & 0.6413\\
Indo-European  & 0.7367 & 0.7752 & 0.7337 &  \textbf{0.7812}  & 0.715 & 0.7126 & 0.7316 & 0.6583\\
Japanese dialects  & 0.893 & 0.9031 & 0.8943 & 0.9051 & 0.9006 &  \textbf{0.9083}  & 0.8875 & 
0.8699\\
Kadai  & 0.7581 & 0.8175 & 0.8139 &  \textbf{0.8309}  & 0.8 & 0.8159 &  --  & 0.7647\\
Kamasau  & 0.9561 &  \textbf{0.9850}  & 0.9543 & 0.9823 & 0.9605 & 0.9674 &  --  & 0.9479\\
Lolo-Burmese  & 0.6469 & 0.713 & 0.7862 & 0.7805 & 0.7846 & \textbf{0.8218} &  
--  &  0.8027 \\
Mayan  &  \textbf{0.8198}  & 0.7798 & 0.6958 & 0.8074 & 0.6804 & 0.6797 & 0.7931 & 0.627\\
Miao-Yao  &  0.6412  & 0.7003 & 0.7679 & 0.7801 & 0.7411 & 0.7879 &  --  & 
\textbf{0.8426}\\
Mixe-Zoque  & 0.9055 & 0.9149 & 0.8528 &  \textbf{0.9209}  & 0.8521 & 0.8599 & 0.8656 & 0.8298\\
Mon-Khmer  & 0.7883 & 0.8209 & 0.7054 &  \textbf{0.8302}  & 0.6921 & 0.7008 & 0.7925 & 0.6472\\
ObUgrian  & 0.8623 & 0.911 & 0.8987 &  \textbf{0.9214}  & 0.8951 & 0.8874 & 0.8837 & 0.8826\\
Tujia  & 0.8882 & 0.9091 & 0.9018 &  \textbf{0.9105}  & 0.895 & 0.9027 & 0.8905 & 
0.8757\\\hline\hline
Average  & 0.8044 & 0.8289 & 0.7971 &  \textbf{0.8415}  & 0.7901 & 0.7955 & 0.8185 & 0.7548\\\hline
    \end{tabular}%
}
\caption{The B-cubed F-scores of different models on sixteen language groups. The last row 
reports the average of the B-cubed F-scores across all the datasets. The numbers in \textbf{bold} 
show the highest scores across columns.}
\label{tab:outfam}%
\end{table*}%

\subsection{Out-of-family training}
In this experiment, we train our PHMM and PMI systems on wordlists from the 
ASJP database belonging to families other than those language groups present in table 
\ref{tab:chardataset}. We made sure that there is no overlap between the languages present in test 
dataset and the training dataset. We extracted a list of probable cognates and trained our
PMI and PHMM models on the list of probable cognates. We trained all the batch and online systems 
on $1151178$ word pairs. The results of our experiments are given in table \ref{tab:outfam}. We 
report the InfoMap clustering results for a threshold of $0.5$ for all the systems. We expect 
LexStat to perform better in the case of Chinese since LexStat handles tones internally whereas, 
the ASJP representation does not handle tones. In the case of online systems, we report the best 
results for $m, \alpha$. Following \citet{list2014investigating}, we do not report LexStat 
results for the language groups which have word lists shorter than $100$ meanings.

\begin{table*}[!ht]
\centering
\small
 \begin{tabular}{l|rl|rl}\hline\hline  
 & \multicolumn{2}{c|}{PMI} & \multicolumn{2}{c}{PHMM}\\
 \cline{2-3}\cline{4-5}
Family & $m$ & $\alpha$ & $m$ & $\alpha$\\\hline
Austronesian & 64 & 0.75 & 32 & 0.5\\
Afrasian & 256 & 0.65 & 32 & 0.8\\
Bai dialects & 8192 & 0.75 & 32 & 0.55\\
Chinese dialects & 128 & 0.95 & 512 & 0.6\\
Huon & 32 & 1 & 32 & 0.65\\
Indo-European & 512 & 0.55 & 1024 & 0.5\\
Japanese dialects & 512 & 0.55 & 32 & 0.6\\
Kadai & 2048 & 0.7 & 32 & 0.7\\
Kamasau & 512 & 0.5 & 128 & 0.55\\
Lolo-Burmese & 16384 & 0.5 & 32 & 0.75\\
Mayan & 64 & 0.5 & 32 & 0.55\\
Miao-Yao & 8192 & 0.95 & 128 & 0.7\\
Mixe-Zoque & 256 & 0.7 & 32 & 0.7\\
Mon-Khmer & 256 & 0.7 & 32 & 0.5\\
ObUgrian & 512 & 0.75 & 32768 & 0.5\\
Tujia & 1024 & 0.65 & 32 & 0.5\\\hline\hline
 \end{tabular}
\caption{Best settings of $m$ and $\alpha$ for Online variants of PMI and PHMM.}
\label{tab:mandalpha}
\end{table*}

The Online PMI performs better than the rest of the systems at nine out of the sixteen families. On 
an average, the Online PMI system ranks the best followed by PMI-LANG and LexStat 
system. ALINE performs the best on Miao-Yao language group. 
The Online PMI system perform better than the Batch PMI on all the datasets. As expected, the 
LexStat system performs the best on Chinese dialect dataset. Surprisingly, despite its complexity 
the PHMM systems do not perform as well as the simpler PMI systems.

Now, we will comment on the results of Austronesian and Indo-European language families. 
\citet{greenhill2011levenshtein} applied Levenshtein distance for the classification of 
Austronesian languages and argued that Levenshtein distance does not perform well at the task of 
detecting language relationships. Our experiment shows that Levenshtein distance comes close to 
LexStat in the case of Austronesian language family. Both PMI-LANG and Online 
PMI are two 
points better than Levenshtein distance at the task of cognate identification.

The results are much clearer in the case of Indo-European language family. The PMI-LANG and Online 
PMI systems perform better than rest of the systems. Levenshtein distance performs better than 
LexStat for the Indo-European language family. On an average, ALINE shows the lowest performance of 
all the systems.

\begin{figure*}[!ht]
\centering
 \begin{subfigure}[b]{0.45\textwidth} 
\includegraphics[height=\textwidth]{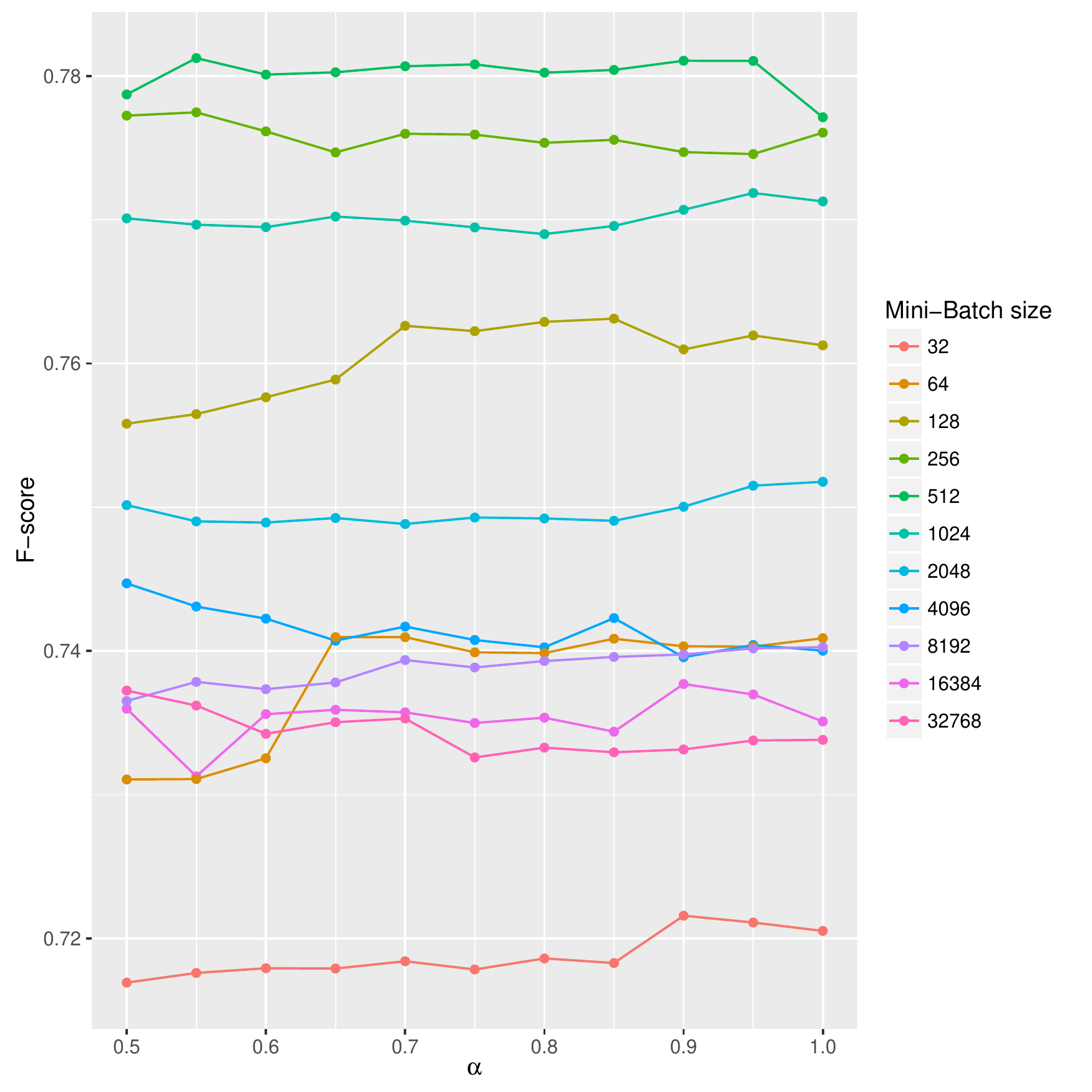}
\caption{Indo-European}
\label{fig:globalielex}
 \end{subfigure}
~
\begin{subfigure}[b]{0.45\textwidth} 
\includegraphics[height=\textwidth]{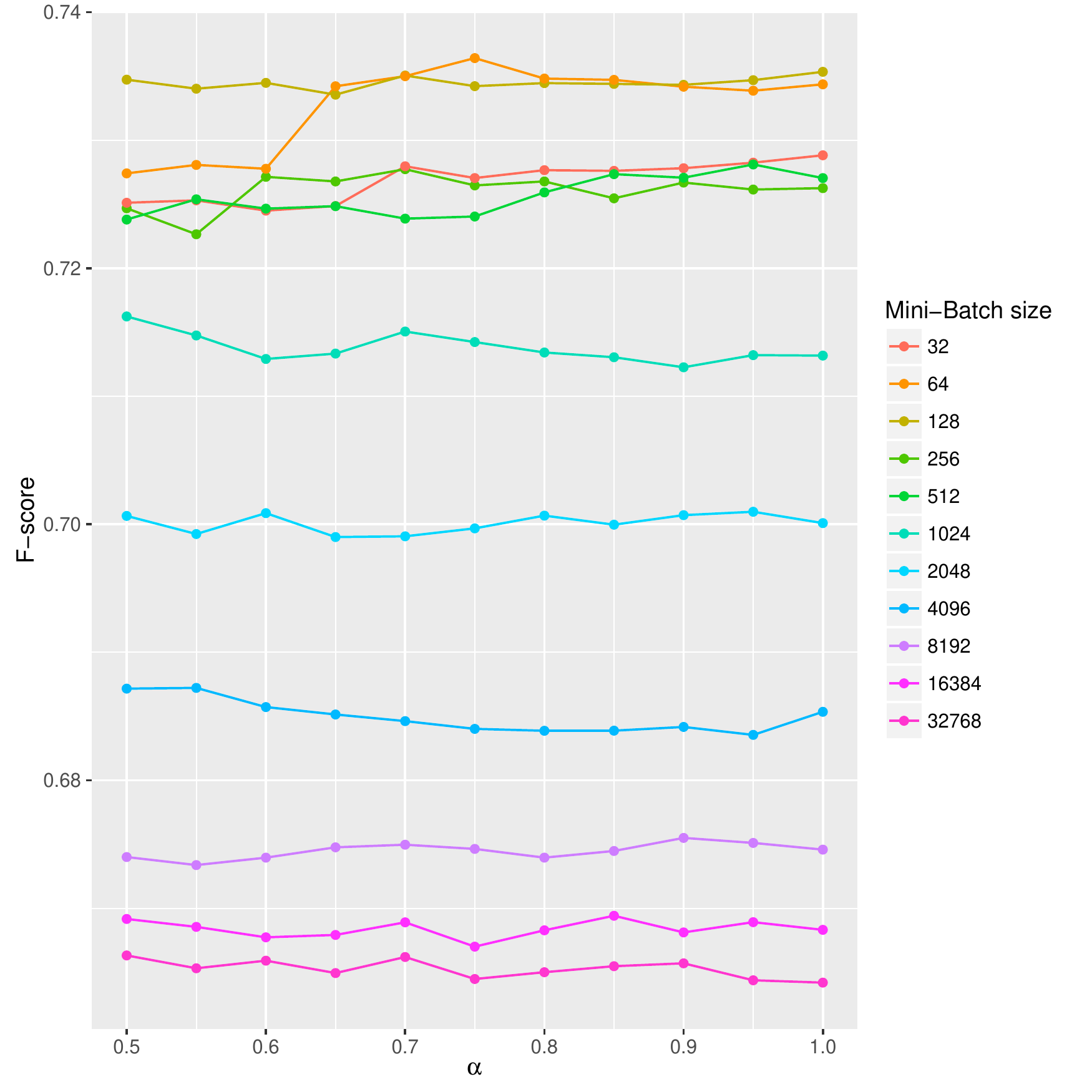}
\caption{Austronesian}
\label{fig:globalabvd}
 \end{subfigure}
\caption{Plots of $m$ and $\alpha$ against B-cubed F-scores for out-of-family 
training.}
 \label{fig:globalplots}
\end{figure*}

We report the corresponding setting of $m, \alpha$ for all the online systems in table 
\ref{tab:mandalpha}. The value of $m$ is quite variable across language 
families whereas, $\alpha$ tends to be in the range of $0.5-0.75$. We 
investigate the effect of $m$ and $\alpha$ for 
Indo-European and Austronesian languages by plotting the results of Online PMI system in figures 
\ref{fig:globalplots}. The B-cubed F-scores are stable across the range of 
$\alpha$ but show variable results for value of $m$. The top-3 F-scores for Indo-European are at 
$m=256, 512, 1024$ and at $m=64, 128, 256$ for Austronesian language family. These results suggest 
that the online training helps cognate clustering than the batch training. The 
plots (cf. figure \ref{fig:globalplots}) suggest that 
small batch size improves the performance whereas a large batch size (eg., 32768) hurts the 
performance on Indo-European and Austronesian language families.

\subsection{Within-family training}
\begin{table*}[!t]
  \centering
\scalebox{0.65}{
\begin{tabular}{l|c|ccc|ccc|cc}
\hline
 \multirow{2}{*}{Family} & \multirow{2}{*}{Training word pairs} & 
\multicolumn{3}{|c|}{Online PHMM} & \multicolumn{3}{|c|}{Online 
PMI} & \multirow{2}{*}{Batch PHMM} & \multirow{2}{*}{Batch PMI}\\
\cline{3-5}\cline{6-8}
 & & $m$ & $\alpha$ & F-score & $m$ & $\alpha$ & F-score &  & \\\hline\hline
ASJP Indo-European & 380769 & 128 & 0.60 & 0.7646 & 4096 & 0.60 & 
\textbf{0.7868} & 0.7656 & 0.7704\\
Indo-European & 25386 & 64 & 0.50 & 0.7901 & 1024 & 0.85 & \textbf{0.7971} & 
0.7797 & 0.7914\\\hline\hline
ASJP Mayan & 91665 & 256 & 0.55 & 0.7765 & 128 & 0.90 & \textbf{0.8250} & 
0.7814 & 0.7677\\
Mayan & 11889 & 32 & 0.55 & 0.7952 & 64 & 0.70 & \textbf{0.7997} & 0.7888 & 
0.7544\\\hline\hline
ASJP Austronesian & 1000000 & 32 & 0.65 & 0.6190 & 128 & 0.80 & \textbf{0.7453} 
& 0.6239 & 0.6429\\
Austronesian & 84311 & 32 & 0.5 & 0.6709 & 128 & 0.80 & \textbf{0.7460} & 
0.6517 & 0.6509\\\hline\hline
\end{tabular}
}
\caption{The results of training the PMI and PHMM systems on the ASJP 40 word lists and the 
full word lists of Indo-European, Mayan, and Austronesian.}
\label{tab:withinfam}
\end{table*}
In this experiment, we train our PMI and PHMM systems on three largest language families in our 
dataset: Mayan, Indo-European, and Austronesian language families. We train our 
systems on word pairs extracted from two different sources.
\begin{enumerate}
 \item The ASJP database has $40$-length word lists for more 
languages ($\sim$ 3 times) than the languages in cognate databases of Mayan, 
Indo-European, and Austronesian language 
families. The database allows us to access more word pairs than any other database in existence.
\item We extract list of probable cognate pairs from the IELex, ABVD, and Mayan language databases.
\end{enumerate}

The motivation behind these experiments is to investigate the performance of PMI and PHMM systems 
when trained on the word lists belonging to the same language family but compiled by different 
group of annotators. A successful experiment indicates that this approach of training a PMI matrix 
on ASJP $40$ word lists can be applied to language families that have longer word lists but no 
cognate judgments. The number of training word pairs and the results of our experiments are given 
in table \ref{tab:withinfam}.

The Online variants perform better than the batch systems across all the language families and 
settings. Online PMI performs the best across all the language families than the 
Batch PMI. Online PMI trained on ASJP word lists of a language family show close 
performance to an Online PMI system trained within the language family in the 
case of Indo-European and Austronesian language families. The performance of 
batch PMI system comes close to the Online PMI system in the case of 
Indo-European but falls behind in the case of other language families. Training 
the online system on ASJP word lists improves the performance in the case of 
Mayan language family. This performance is not observed in the case of 
Indo-European and Austronesian language families. The reason for 
this could be due to the source of origin of the datasets.

The batch PMI/PHMM systems perform better than LexStat on Indo-European and Mayan language 
families. The Online PHMM system comes close in performance to Online PMI system in the case of 
Indo-European and Mayan language families. PHMM systems how the lowest performance on Austronesian 
language family. Except for Indo-European, the best batch sizes for online PMI system are 
small and are typically $\le 256$. 

\section{Discussion}\label{sec:erranal}
In this section, we discuss the effects of various parameters on our results.

\subsection{Effect of $m$ and $\alpha$} Throughout our experiments, we observe that 
low 
minibatch size gives better results than a large minibatch size. We also observe that a 
intermediary value of $\alpha$ is usually sufficient for obtaining the best results.

Figure \ref{fig:globalplots} shows that small values of $m$ yields stable F-scores across the range 
of $\alpha$. Small values of $m$ typically gives better results than larger values of $\alpha$. In 
contrast to other NLP tasks that require large $m$ and smaller $\alpha$, the task of aligning two 
words requires smaller values of $m$. The small value of $m$ implies large number of 
updates which is important for a task where the average sequence length ($\sim 5$) and the average 
number of word pairs are in less than $100,000$. Further, an intermediary value of $\alpha$ 
controls the amount of memory retained at each update.

\subsection{Speed} One advantage of our online systems (either PMI or PHMM) is that the training 
time 
is typically in the range of $10$ minutes on a single thread of i7-6700 processor. In the 
case of PHMM, online training speeds up the convergence and yields, typically, better results than 
the batch variant. In comparison, the PMI-LANG system takes days to train. Finally, our results 
show 
that the online algorithm can yield better performance than LexStat. LexStat and PHMM take more 
than 
5 hours to test on the language subset of the Austronesian language family. In contrast, PMI (both 
online and batch) takes less than $10$ minutes for each value of $m, \alpha$ in the case of 
out-of-family training. We also observe that $5$ 
scans over the full data was sufficient for convergence.

\subsection{Analyzing PHMM's performance}

Although PHMMs are the most complex among 
the tested models, the performance of these models is not as good as the 
conceptually simpler PMI models. This lack of performance could be due to the 
characteristics of the PHMM. The transition probability from the begin state to 
the match or gap states is the same as the transition probability from the match 
state to either gap state or itself (figure \ref{fig:PairHMMkondrak}). 
Although desirable for biological purposes, this poses a big problem for 
linguistic applications. To start an alignment with a match is more likely than 
to start with a gap.\footnote{$1-2\delta\-\tau_M$ 
is larger than $\delta$ in all models (c.f. figure \ref{fig:PairHMMkondrak}).} 
\begin{figure}[!ht]
\centering
 \includegraphics[scale=0.4]{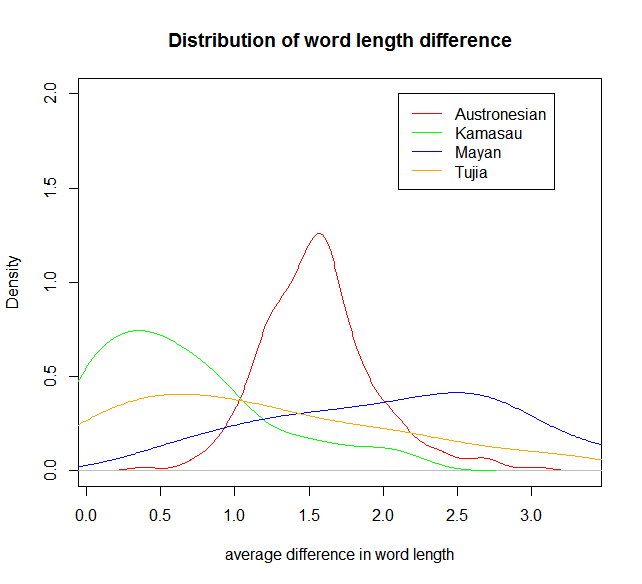}
 \caption{Distribution of average of word length differences across concepts.}
 \label{fig:wldst}
\end{figure}
Therefore, the alignments 
generated by PHHMs are more likely to show gaps at the end of the string than in 
the beginning. This results in problems for data sets where word length differ 
a lot. The PHMM performs the worst for those datasets that show a huge 
difference in the word length. On the other hand, for Kamasau and Tujia -- the 
two datasets with the best performance -- the difference in word length is much 
less pronounced (cf. figure \ref{fig:wldst}).

Based on the results of these experiments, we propose that training the PMI-based segment scores in 
an online fashion and supplied to InfoMap clustering could yield reliable 
cognate judgments.

\section{Conclusion}\label{sec:concl}
In this paper, we evaluated the performance of various sequence alignment algorithms -- both 
learned and linguistically designed -- for the 
task of cognate detection across different language families. We find that training PMI and PHMM in 
an online fashion speeds up convergence and yields comparable or better results than the batch 
variant and the state-of-the-art LexStat system. Online PMI system shows the 
best performance across different language families. In conclusion, PMI systems can be 
trained faster in an online fashion and yield better accuracies than the current state-of-the-art 
systems.

\bibliographystyle{plainnat}
\bibliography{literature}

\begin{thebibliography}{33}
\providecommand{\natexlab}[1]{#1}
\providecommand{\url}[1]{\texttt{#1}}
\expandafter\ifx\csname urlstyle\endcsname\relax
  \providecommand{\doi}[1]{doi: #1}\else
  \providecommand{\doi}{doi: \begingroup \urlstyle{rm}\Url}\fi

\bibitem[Amig{\'o} et~al.(2009)Amig{\'o}, Gonzalo, Artiles, and
  Verdejo]{amigo2009comparison}
Enrique Amig{\'o}, Julio Gonzalo, Javier Artiles, and Felisa Verdejo.
\newblock A comparison of extrinsic clustering evaluation metrics based on
  formal constraints.
\newblock \emph{Information retrieval}, 12\penalty0 (4):\penalty0 461--486,
  2009.

\bibitem[Bouchard-C{\^o}t{\'e} et~al.(2013)Bouchard-C{\^o}t{\'e}, Hall,
  Griffiths, and Klein]{bouchardcote2013}
Alexandre Bouchard-C{\^o}t{\'e}, David Hall, Thomas~L. Griffiths, and Dan
  Klein.
\newblock Automated reconstruction of ancient languages using probabilistic
  models of sound change.
\newblock \emph{Proceedings of the National Academy of Sciences}, 110\penalty0
  (11):\penalty0 4224--4229, 2013.
\newblock \doi{10.1073/pnas.1204678110}.
\newblock URL
  \url{http://www.pnas.org/content/early/2013/02/05/1204678110.abstract}.

\bibitem[Bouckaert et~al.(2012)Bouckaert, Lemey, Dunn, Greenhill, Alekseyenko,
  Drummond, Gray, Suchard, and Atkinson]{bouckaert2012mapping}
Remco Bouckaert, Philippe Lemey, Michael Dunn, Simon~J. Greenhill, Alexander~V.
  Alekseyenko, Alexei~J. Drummond, Russell~D. Gray, Marc~A. Suchard, and
  Quentin~D. Atkinson.
\newblock Mapping the origins and expansion of the {I}ndo-{E}uropean language
  family.
\newblock \emph{Science}, 337\penalty0 (6097):\penalty0 957--960, 2012.

\bibitem[Brown et~al.(2013)Brown, Holman, and Wichmann]{brown2013sound}
Cecil~H. Brown, Eric~W. Holman, and S{\o}ren Wichmann.
\newblock Sound correspondences in the world's languages.
\newblock \emph{Language}, 89\penalty0 (1):\penalty0 4--29, 2013.

\bibitem[Church and Hanks(1990)]{Church:90}
Kenneth~Ward Church and Patrick Hanks.
\newblock Word association norms, mutual information, and lexicography.
\newblock \emph{Computational Linguistics}, 16\penalty0 (1):\penalty0 22--29,
  1990.
\newblock ISSN 0891-2017.

\bibitem[Downey et~al.(2008)Downey, Hallmark, Cox, Norquest, and
  Lansing]{downey2008computational}
Sean~S Downey, Brian Hallmark, Murray~P Cox, Peter Norquest, and J~Stephen
  Lansing.
\newblock Computational feature-sensitive reconstruction of language
  relationships: Developing the aline distance for comparative historical
  linguistic reconstruction.
\newblock \emph{Journal of Quantitative Linguistics}, 15\penalty0 (4):\penalty0
  340--369, 2008.

\bibitem[Durbin et~al.(2001)Durbin, Eddy, Krogh, and Mitchison]{Durbinetal}
Richard Durbin, Sean~R Eddy, Anders Krogh, and Graeme Mitchison.
\newblock \emph{Biological sequence analysis: probabilistic models of proteins
  and nucleic acids}.
\newblock Cambridge Univ. Press, repr. edition, 2001.

\bibitem[Dyen et~al.(1992)Dyen, Kruskal, and Black]{dyen1992indoeuropean}
Isidore Dyen, Joseph~B. Kruskal, and Paul Black.
\newblock An {I}ndo-{E}uropean classification: A lexicostatistical experiment.
\newblock \emph{Transactions of the American Philosophical Society},
  82\penalty0 (5):\penalty0 1--132, 1992.

\bibitem[Gray and Atkinson(2003)]{gray2003language}
Russell~D Gray and Quentin~D Atkinson.
\newblock Language-tree divergence times support the anatolian theory of
  indo-european origin.
\newblock \emph{Nature}, 426\penalty0 (6965):\penalty0 435--439, 2003.

\bibitem[Greenhill(2011)]{greenhill2011levenshtein}
Simon~J Greenhill.
\newblock Levenshtein distances fail to identify language relationships
  accurately.
\newblock \emph{Computational Linguistics}, 37\penalty0 (4):\penalty0 689--698,
  2011.

\bibitem[Greenhill and Gray(2009)]{greenhill2009austronesian}
Simon~J. Greenhill and Russell~D. Gray.
\newblock {Austronesian language phylogenies: Myths and misconceptions about
  Bayesian computational methods}.
\newblock \emph{Austronesian Historical Linguistics and Culture History: A
  Festschrift for Robert Blust}, pages 375--397, 2009.

\bibitem[Hauer and Kondrak(2011)]{Hauer2011}
Bradley Hauer and Grzegorz Kondrak.
\newblock Clustering semantically equivalent words into cognate sets in
  multilingual lists.
\newblock In \emph{Proceedings of the 5th International Joint Conference on
  Natural Language Processing}, pages 865--873, 2011.

\bibitem[Huff and Lonsdale(2011)]{huff2011positing}
Paul Huff and Deryle Lonsdale.
\newblock Positing language relationships using aline.
\newblock \emph{Language Dynamics and Change}, 1\penalty0 (1):\penalty0
  128--162, 2011.

\bibitem[J{\"a}ger(2013)]{jager2013phylogenetic}
Gerhard J{\"a}ger.
\newblock Phylogenetic inference from word lists using weighted alignment with
  empirically determined weights.
\newblock \emph{Language Dynamics and Change}, 3\penalty0 (2):\penalty0
  245--291, 2013.

\bibitem[J\"ager(2015)]{Jaeger2015}
Gerhard J\"ager.
\newblock Support for linguistic macrofamilies from weighted sequence
  alignment.
\newblock \emph{Proceedings of the National Academy of Sciences}, 112\penalty0
  (41):\penalty0 12752--12757, 2015.
\newblock \doi{{10.1073/pnas.1500331112}}.

\bibitem[Kondrak(2000)]{kondrak2000new}
Grzegorz Kondrak.
\newblock A new algorithm for the alignment of phonetic sequences.
\newblock In \emph{Proceedings of the 1st North American chapter of the
  Association for Computational Linguistics conference}, pages 288--295.
  Association for Computational Linguistics, 2000.

\bibitem[Ladefoged(1975)]{Ladefoged1975}
Peter Ladefoged.
\newblock \emph{{A course in phonetics}}.
\newblock Hardcourt Brace Jovanovich Inc. NY, 1975.

\bibitem[Levenshtein(1966)]{Levenshtein1966}
V.~I. Levenshtein.
\newblock Binary codes capable of correcting deletions, insertions, and
  reversals.
\newblock \emph{Soviet Physics Doklady}, 10\penalty0 (8):\penalty0 707--710,
  1966.

\bibitem[Liang and Klein(2009)]{LiangKlein2009}
Percy Liang and Dan Klein.
\newblock Online em for unsupervised models.
\newblock In \emph{Proceedings of Human Language Technologies: The 2009 Annual
  Conference of the North American Chapter of the Association for Computational
  Linguistics}, NAACL '09, pages 611--619, Stroudsburg, PA, USA, 2009.
  Association for Computational Linguistics.
\newblock ISBN 978-1-932432-41-1.
\newblock URL \url{http://dl.acm.org/citation.cfm?id=1620754.1620843}.

\bibitem[List(2012)]{list2012lexstat}
Johann-Mattis List.
\newblock Lexstat: Automatic detection of cognates in multilingual wordlists.
\newblock In \emph{Proceedings of the EACL 2012 Joint Workshop of LINGVIS \&
  UNCLH}, pages 117--125. Association for Computational Linguistics, 2012.

\bibitem[List(2014{\natexlab{a}})]{List2014d}
Johann-Mattis List.
\newblock \emph{Sequence comparison in historical linguistics}.
\newblock Düsseldorf University Press, Düsseldorf, 2014{\natexlab{a}}.
\newblock URL \url{http://sequencecomparison.github.io/}.

\bibitem[List(2014{\natexlab{b}})]{list2014investigating}
Johann-Mattis List.
\newblock Investigating the impact of sample size on cognate detection.
\newblock \emph{Journal of Language Relationship}, 11:\penalty0 91--101,
  2014{\natexlab{b}}.

\bibitem[List and Forkel(2016)]{List2016e}
Johann-Mattis List and Robert Forkel.
\newblock Lingpy. a python library for historical linguistics, 2016.
\newblock URL \url{http://lingpy.org}.

\bibitem[List et~al.(2016)List, Lopez, and
  Bapteste]{list-lopez-bapteste:2016:P16-2}
Johann-Mattis List, Philippe Lopez, and Eric Bapteste.
\newblock Using sequence similarity networks to identify partial cognates in
  multilingual wordlists.
\newblock In \emph{Proceedings of the 54th Annual Meeting of the Association
  for Computational Linguistics (Volume 2: Short Papers)}, pages 599--605,
  Berlin, Germany, August 2016. Association for Computational Linguistics.
\newblock URL \url{http://anthology.aclweb.org/P16-2097}.

\bibitem[Mackay and Kondrak(2005)]{MackayKondrak2005}
Wesley Mackay and Grzegorz Kondrak.
\newblock {Computing word similarity and identifying cognates with pair hidden
  Markov models}.
\newblock CONLL '05, pages 40--47, Stroudsburg, PA, USA, June 2005. Association
  for Computational Linguistics.

\bibitem[Needleman and Wunsch(1970)]{NW}
Saul~B. Needleman and Christian~D. Wunsch.
\newblock A general method applicable to the search for similarities in the
  amino acid sequence of two proteins.
\newblock \emph{Journal of Molecular Biology}, 48\penalty0 (3):\penalty0 10,
  1970.

\bibitem[Newman and Girvan(2004)]{NewmanGirvan2004}
Mark~EJ Newman and Michelle Girvan.
\newblock Finding and evaluating community structure in networks.
\newblock \emph{Phys. Rev. E}, 69:\penalty0 026113, Feb 2004.
\newblock \doi{10.1103/PhysRevE.69.026113}.
\newblock URL \url{http://link.aps.org/doi/10.1103/PhysRevE.69.026113}.

\bibitem[Rama(2015)]{rama2015automatic}
Taraka Rama.
\newblock Automatic cognate identification with gap-weighted string
  subsequences.
\newblock In \emph{Proceedings of the 2015 Conference of the North American
  Chapter of the Association for Computational Linguistics: Human Language
  Technologies.}, pages 1227--1231, 2015.

\bibitem[Rama and Borin(2015)]{rama2013bchap}
Taraka Rama and Lars Borin.
\newblock Comparative evaluation of string similarity measures for automatic
  language classification.
\newblock In Ján Mačutek and George~K. Mikros, editors, \emph{Sequences in
  Language and Text}, pages 203--231. Walter de Gruyter, 2015.

\bibitem[Rosenberg and Hirschberg(2007)]{rosenberg2007v}
Andrew Rosenberg and Julia Hirschberg.
\newblock V-measure: A conditional entropy-based external cluster evaluation
  measure.
\newblock In \emph{EMNLP-CoNLL}, volume~7, pages 410--420, 2007.

\bibitem[Rosvall and Bergstrom(2008)]{Rosvalletal2008}
Martin Rosvall and Carl~T. Bergstrom.
\newblock Maps of random walks on complex networks reveal community structure.
\newblock \emph{Proceedings of the National Academy of Sciences}, 105\penalty0
  (4):\penalty0 1118--1123, 2008.
\newblock \doi{10.1073/pnas.0706851105}.
\newblock URL \url{http://www.pnas.org/content/105/4/1118.abstract}.

\bibitem[Wichmann and Holman(2013)]{wichmann2013languages}
S{\o}ren Wichmann and Eric~W Holman.
\newblock Languages with longer words have more lexical change.
\newblock In \emph{Approaches to Measuring Linguistic Differences}, pages
  249--281. Mouton de Gruyter, 2013.

\bibitem[Wieling et~al.(2007)Wieling, Leinonen, and
  Nerbonne]{WielingNerbonne2007}
Martijn Wieling, Therese Leinonen, and John Nerbonne.
\newblock {Inducing sound segment differences using Pair Hidden Markov Models}.
\newblock pages 48--56. Association for Computational Linguistics, 2007.

\end{thebibliography}

\end{document}